\documentclass[letterpaper, 10 pt, conference]{ieeeconf}  

\IEEEoverridecommandlockouts                              

\usepackage{graphics} 
\usepackage{epsfig} 
\usepackage{mathptmx} 
\usepackage{times} 
\usepackage{amsmath} 
\usepackage{amssymb}  
\usepackage{multirow}
\usepackage{amsfonts}
\usepackage{booktabs}
\usepackage[table,xcdraw]{xcolor}
\usepackage{cite}
\usepackage{url}
\usepackage{threeparttable}
\usepackage{bigdelim}
\usepackage{indentfirst}
\usepackage{colortbl}
\usepackage{bbding}
\usepackage[pagebackref=true,breaklinks=true, colorlinks, bookmarks=false]{hyperref}
\definecolor{car}{rgb}{0.39215686, 0.58823529, 0.96078431}
\definecolor{bicycle}{rgb}{0.39215686, 0.90196078, 0.96078431}
\definecolor{motorcycle}{rgb}{0.11764706, 0.23529412, 0.58823529}
\definecolor{truck}{rgb}{0.31372549, 0.11764706, 0.70588235}
\definecolor{other-vehicle}{rgb}{0.39215686, 0.31372549, 0.98039216}
\definecolor{person}{rgb}{1.        , 0.11764706, 0.11764706}
\definecolor{bicyclist}{rgb}{1.        , 0.15686275, 0.78431373}
\definecolor{motorcyclist}{rgb}{0.58823529, 0.11764706, 0.35294118}
\definecolor{road}{rgb}{1.        , 0.        , 1.        }
\definecolor{parking}{rgb}{1.        , 0.58823529, 1.        }
\definecolor{sidewalk}{rgb}{0.29411765, 0.        , 0.29411765}
\definecolor{other-ground}{rgb}{0.68627451, 0.        , 0.29411765}
\definecolor{building}{rgb}{1.        , 0.78431373, 0.        }
\definecolor{fence}{rgb}{1.        , 0.47058824, 0.19607843}
\definecolor{vegetation}{rgb}{0.        , 0.68627451, 0.        }
\definecolor{trunk}{rgb}{0.52941176, 0.23529412, 0.        }
\definecolor{terrain}{rgb}{0.58823529, 0.94117647, 0.31372549}
\definecolor{pole}{rgb}{1.        , 0.94117647, 0.58823529}
\definecolor{traffic-sign}{rgb}{1.        , 0.        , 0.    }    

\title{\LARGE \bf
SSC-RS: Elevate LiDAR Semantic Scene Completion with \\ 
 Representation Separation and BEV Fusion
}

\author{
Jianbiao Mei$^{1}$, Yu Yang$^{1}$, Mengmeng Wang$^{1}$, Tianxin Huang$^{1}$, Xuemeng Yang$^{2}$ and Yong Liu$^{1,\dag}$
\thanks{
$^{1}$The authors are with the Institute of Cyber-Systems and Control, Zhejiang University, Hangzhou, 310027, China. (Yong Liu$\dag$ is the corresponding author, email: yongliu@iipc.zju.edu.cn)}
\thanks{
$^{2}$The authors are with the Shanghai Artificial Intelligence Laboratory, Shanghai, China.
}}%

\begin{document}

\maketitle
\thispagestyle{empty}
\pagestyle{empty}

\begin{abstract}
Semantic scene completion (SSC) jointly predicts the semantics and geometry of the entire 3D scene, which plays an essential role in 3D scene understanding for autonomous driving systems. SSC has achieved rapid progress with the help of semantic context in segmentation. However, how to effectively exploit the relationships between the semantic context in semantic segmentation and geometric structure in scene completion remains under exploration. In this paper, we propose to solve outdoor SSC from the perspective of representation separation and BEV fusion. Specifically, we present the network, named SSC-RS, which uses separate branches with deep supervision to explicitly disentangle the learning procedure of the semantic and geometric representations. And a BEV fusion network equipped with the proposed Adaptive Representation Fusion (ARF) module is presented to aggregate the multi-scale features effectively and efficiently. Due to the low computational burden and powerful representation ability, our model has good generality while running in real-time. Extensive experiments on SemanticKITTI demonstrate our SSC-RS achieves state-of-the-art performance. Code is available at \url{https://github.com/Jieqianyu/SSC-RS.git}.
\end{abstract}

\section{Introduction}
In recent years, 3D scene understanding, one of the most important functions of perception systems in autonomous driving, has attracted extensive studies and achieved rapid progress. When working with large-scale outdoor scene understanding, Semantic Scene Completion (SSC) aims to predict the semantic occupancy of each voxel of the entire 3D scene from the sparse LiDAR scans, including the completion of certain regions. Due to the ability to recover geometric structure, SSC can facilitate further applications like 3D object detection, which usually suffer from the sparsity and incompleteness (caused by occlusions or far distance from sensors) of the LiDAR point cloud. However, it's challenging to precisely estimate the semantics and geometry of the whole 3D real-world scene from partial observations due to the complex outdoor scenarios such as various shapes/sizes and occlusions.

Following the pioneering work SSCNet \cite{song2017semantic}, some existing outdoor SSC methods \cite{zou2021up, roldao2020lmscnet} exploit a single U-Net network, e.g., a heavy dense 3D convolution network to predict semantics and geometry jointly. However, they usually involve unnecessary calculations and extra memory and computation overhead, especially when the input voxel resolution is large since there are lots of empty voxels in the 3D scene. 

\begin{figure}[t]
\centering
	\includegraphics[width=0.48\textwidth]{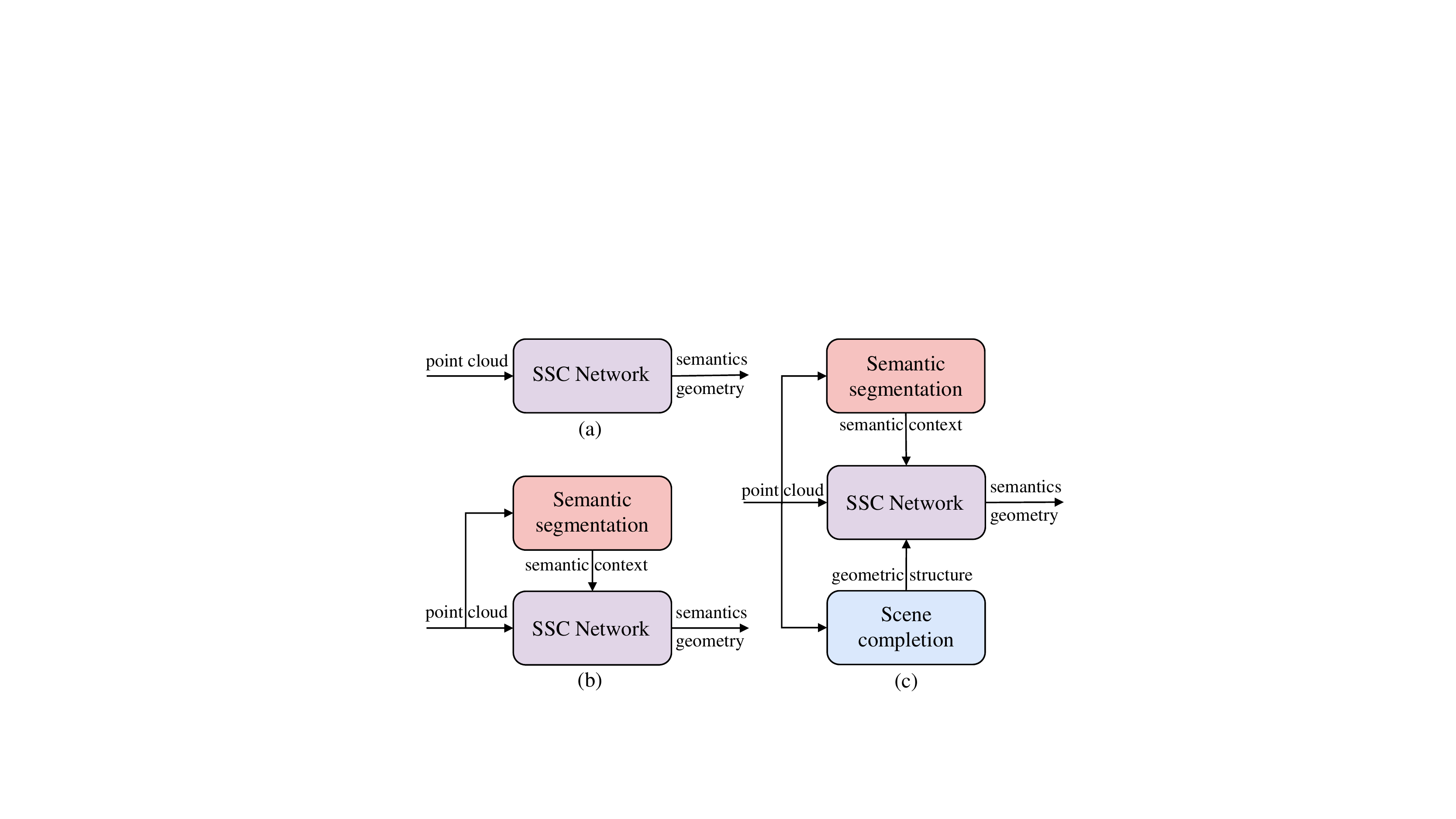}
	\caption{(a) Consider the semantic context and geometric structure in a hybrid manner. (b) Consider the semantic context and geometric structure in a semi-hybrid manner. (c) Disentangling the learning procedure of semantic context and geometric structure explicitly.}
	\label{fig:intro}
\end{figure}

On the other way, some methods \cite{yan2021sparse, yang2021semantic, cheng2021s3cnet, garbade2019two} utilize the semantic information in the segmentation to assist outdoor SSC by combining the semantic completion network with the segmentation network.
We found that most outdoor SSC methods consider the semantic context (semantic representation) and geometry structure (geometric representation) in a hybrid (Fig. \ref{fig:intro} (a)) or semi-hybrid manner (Fig. \ref{fig:intro} (b)). And how to effectively learn semantic/geometric representations and exploit their relationship remains unexplored.

In this paper, we explore the solutions to the outdoor SSC from the perspective of representation separation and BEV (Bird's-Eye View) fusion (Fig. \ref{fig:intro} (c)). We propose to explicitly disentangle the learning procedure of semantic context and geometric structure and fuse them in the BEV, which is demonstrated to be a kind of success in 3D object detection and segmentation \cite{aksoy2019salsanet, zhang2020polarnet, li2022panoptic, lang2019pointpillars, liu2022bevfusion, zhou2021panoptic}. Our main insights are: (1) Semantic context and geometric structure complement each other and are vital for SSC tasks. Recovering the geometry details according to the semantics is easy, and the completed shapes help identify semantic categories. (2) Explicitly disentangling the representations can facilitate and accelerate the learning procedure of semantic context and geometric structure. (3) Compared to dense feature fusion in 3D space, BEV fusion is more convenient and efficient. 

Specifically, we design separate branches, i.e., semantic and completion branches, for semantic/geometric representations according to their intrinsic properties. We also develop a BEV fusion network to aggregate the two types of representations from the two branches. We use a sparse 3D CNN \cite{graham20183d} to encode the multi-level semantic context and a tiny dense 3D CNN to obtain the multi-scale geometric structures. In addition, we apply deep supervision on both branches to facilitate representation learning. Furthermore, to obtain selective cues from semantic/geometric representations and fuse the semantic context and geometric details sufficiently, we propose an Adaptive Representation Fusion (ARF) module in the BEV fusion network. Due to the lower computational burden and more powerful representation ability, our model runs in real-time and has good generality. Experiments on SemanticKITTI dataset (hidden test) show that our approach achieves state-of-the-art performance (rank the $2^{rd}$ by mIoU and $1^{rd}$ in terms of completion metric (IoU) on the public SSC benchmark\footnote{\url{https://codalab.lisn.upsaclay.fr/competitions/7170\#results}}). Our contributions are summarized as follows:

$\bullet$ We develop the SSC-RS network to solve the outdoor SSC problem from the perspective of representation (semantic/geometric) separation and BEV fusion.

$\bullet$ We design two separate branches for multi-level semantic context and multi-scale geometric structures according to their properties. And deep supervision is applied to facilitate the learning procedure.

$\bullet$ We use 2D CNN as the BEV fusion network to aggregate semantic/geometric representations. And an Adaptive Representation Fusion (ARF) module in the BEV fusion network is proposed to fuse the semantic context and geometric details sufficiently.

$\bullet$ Due to the lightweight design and more powerful representation ability, SSC-RS has low decay and achieves state-of-the-art performance on the SemanticKITTI benchmark.

\section{Related Work}
\subsection{Semantic scene completion}
The indoor SSC method has developed rapidly with the emergence of indoor benchmarks such as SUNCG \cite{song2017semantic} and NYU \cite{silberman2012indoor}. 
Existing methods use different types of geometrical inputs cooperated with corresponding network architectures to complete indoor SSC. For example, \cite{song2017semantic, dai2018scancomplete, wang2019forknet} process the depth maps with 3D CNNs end-to-end. \cite{liu2018see, li2019rgbd, liu20203d, li2020anisotropic, li2020attention} take the RGB-D images with 2D-3D CNNs to explore the modality complementarity. \cite{cai2021semantic, cheng2021s3cnet, wang2019learning, li2019depth} encode the truncated signed distance function (TSDF) representations with the volume network architectures. \cite{zhong2020semantic, tang2022not, huang2021rfnet, zhang2021view, rist2021semantic} process the point clouds with the point-based network to achieve semantic scene completion continuously.  

Since SemanticKITTI \cite{behley2019semantickitti} introduces a large-scale outdoor benchmark for SSC tasks, several outdoor SSC methods have emerged. Following the pioneering work SSCNet \cite{song2017semantic}, \cite{zou2021up} exploits a single U-Net framework to process segmentation and completion simultaneously, resulting in extra computation overhead of empty voxels. Some methods \cite{roldao2020lmscnet, zhang2018efficient} use sparse convolutions or introduce 2D CNN to solve the above problem. 
For example, LMSCNet \cite{roldao2020lmscnet} appends a 3D decoder after the lightweight 2D backbone, and Zhang et al. \cite{zhang2018efficient} designs a sparse CNN with dense deconvolution layers.
Some solutions focusing on multi-view fusion \cite{cheng2021s3cnet}, and local implicit functions \cite{rist2021semantic} are also explored.

Besides, some methods \cite{yan2021sparse, yang2021semantic} exploit semantic segmentation to assist SSC. JS3C-Net \cite{yan2021sparse} inserts a semantic segmentation network before SSC, and SSA-SC \cite{yang2021semantic} injects the features from the segmentation branch into the completion branch hierarchically. In this work, we
propose SSC-RS for large-scale outdoor SSC from the perspective of representation separation and BEV fusion. And different from \cite{xu2022casfusionnet}, which designs a cascaded network to implement the complementary between scene completion and semantic segmentation for indoor SSC, our SSC-RS exploits the multi-scale context. And our parallel feature fusion in BEV is more convenient and efficient.

\subsection{BEV perception in segmentation}
BEV perception indicates vision algorithms in the sense of the BEV view representation for autonomous driving \cite{li2022delving}, which has been explored for a variety of tasks such as LiDAR detection \cite{lang2019pointpillars, zhou2022centerformer}, LiDAR segmentation \cite{li2022panoptic, zhou2021panoptic, zhang2020polarnet}, and sensor fusion \cite{liu2022bevfusion, li2022bevformer}. SalsaNet \cite{aksoy2019salsanet} projects point clouds into BEV feature maps, and PolarNet \cite{zhang2020polarnet} proposes a polar BEV representation for semantic segmentation. Panoptic-PHNet \cite{li2022panoptic} exploits BEV features to enhance the segmentation and perform instance grouping in BEV. Panoptic-Polarnet \cite{zhou2021panoptic} uses a polar BEV representation to implement semantic segmentation and class-agnostic instance clustering. In SSC tasks, S3CNet \cite{cheng2021s3cnet} designs a 2D S3CNet to predict the 2D SSC Image in the BEV of the input point cloud. And SSA-SC \cite{yang2021semantic} take the 2D CNN as the semantic completion network to simultaneously predict the semantics and geometry. Similar to SSA-SC, we also use the 2D CNN as the BEV fusion network to efficiently provide the semantic occupancy of the entire 3D scene. And different from SSA-SC, we explicitly disentangle the learning process of semantic context and geometric structure and take the BEV network as a fusion network. Also, we design an adaptive representation module for aggregating the judicious cues in BEV sufficiently.

\begin{figure*}[t]
\centering
	\includegraphics[width=0.9\textwidth]{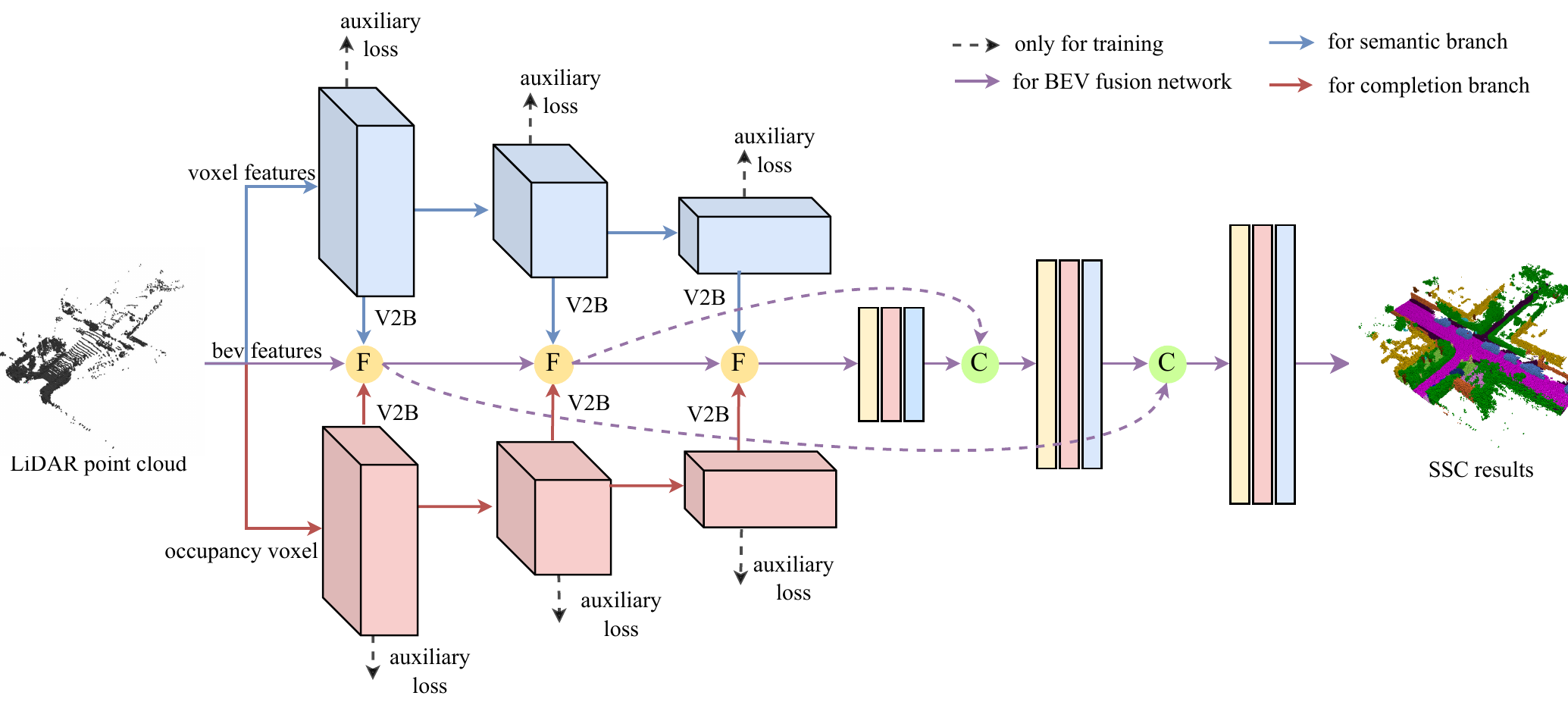}
	\caption{The overview of the proposed SSC-RS. Two branches (semantic/completion branches) are used to learn semantic and geometric representations separately. Both branches are supervised by multi-level auxiliary losses, which will be removed during inference. The multi-scale semantic representations from the semantic branch (blue, sparse 3D CNN) and geometric representations from the completion branch (red, dense 3D CNN) will be fused by the adaptive representation (ARF) module in the BEV fusion network (purple). 'F' denotes the ARF module, and 'C' indicates concatenation along the channel dimension. V2B represents the projection from voxel to BEV.}
	\label{fig:pipeline}
\end{figure*}

\section{Method}
\subsection{Overview}
In this paper, we explore the solutions to LiDAR semantic scene completion from the perspective of representation separation and BEV fusion. Specifically, we design two separate branches to encode semantic and geometric representations, respectively (Sec. \ref{Sec.RS}). Both branches are compact and lightweight. The semantic branch is a stack of 3D sparse convolutions for learning multi-scale semantic context. The completion branch uses several dense 3D convolutions to acquire multi-scale geometry structures from different stages. Based on the representation separation, the BEV fusion network equipped with the proposed ARF module is presented to aggregate informative multi-level features from semantic/completion branches for final semantic scene completion results (Sec. \ref{Sec.BEV}). Fig. \ref{fig:pipeline} illustrates the overall architecture of the proposed SSC-RS.

\subsection{Semantic-completion Representation Separation} \label{Sec.RS}
As discussed above, semantic context and geometry structures are vital cues for semantic scene completion tasks. Thus, according to the inherent properties of these two types of clues, we design separate architectures for learning semantic and geometric representation, respectively.

\begin{figure}[t]
\centering
	\includegraphics[width=0.48\textwidth]{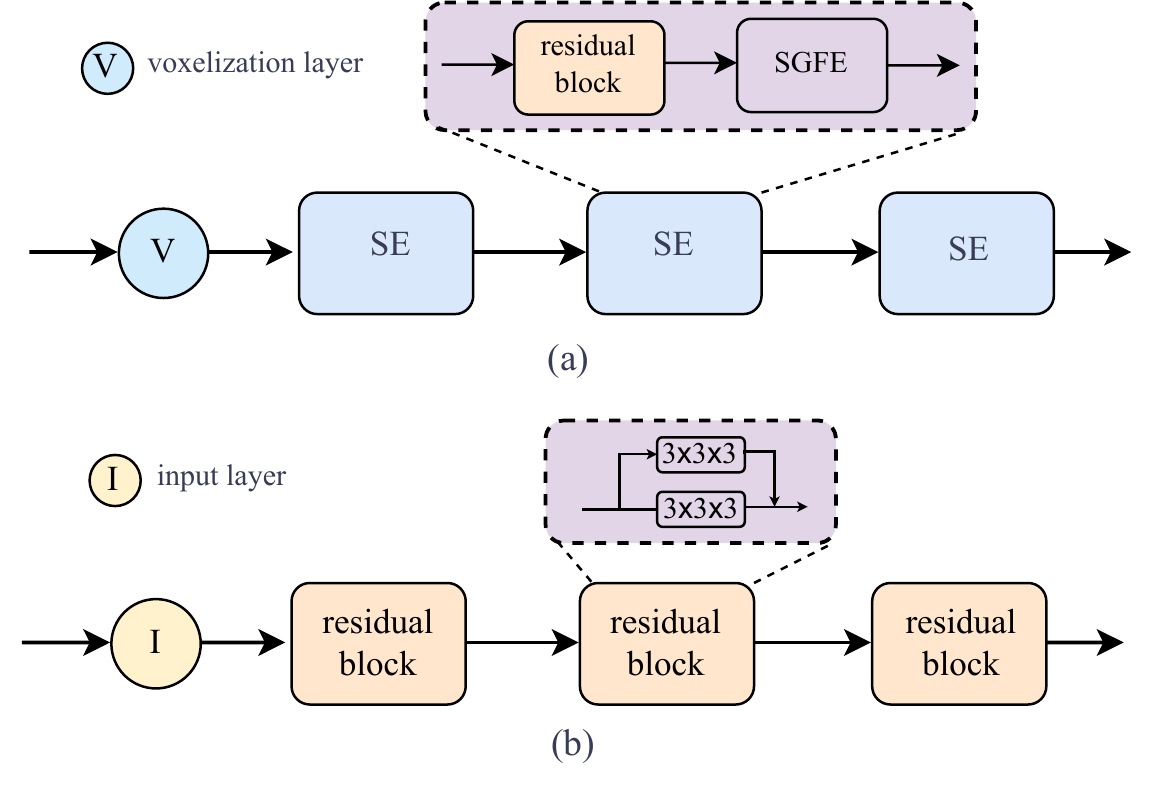}
	\caption{(a) The architecture of the semantic branch. 'SE' denotes the sparse encoder block. (b) The overview of the completion branch. The max-pooling layers are inserted before each residual block.}
	\label{fig:rs}
\end{figure}

\textbf{Semantic Representation:} To encode multi-scale semantic context and improve semantic accuracy, we introduce a compact semantic branch consisting of a voxelization layer and three sparse encoder blocks sharing a similar architecture as shown in Fig. \ref{fig:rs} (a). The voxelization layer takes the point cloud $P\in \mathbb{R}^{N\times3}$ in the range of $[R_x, R_y, R_z]$ as input and outputs sparse voxel features $F_V\in \mathbb{R}^{M\times C}$  with a dense spatial resolution of $L\times W \times H$. It discretizes a point $p_i = (x_i, y_i, z_i)$ to its voxel index $V_i$ through:
\begin{equation}
    V_i = (\lfloor x_i/s \rfloor, \lfloor y_i/s \rfloor, \lfloor z_i/s \rfloor) 
\end{equation} where $s$ is the voxelization resolution and $\lfloor \cdot \rfloor$ is a floor function. Since a occupied voxel could contain multiple points, the voxel features $f_{V_m}$ indexed by $V_m \in \mathbb{Z}^{L \times W \times H}$ are aggregated by:
\begin{equation}
    f_{V_m} = \mathrm{R}_f 
  \left\{ \mathop{\mathrm{A}_f}_{V_p = V_m}(\mathrm{MLP}(f_p)) \right\}
\end{equation} where $\mathrm{A}_f$ is aggregation function (e.g. max function) and $\mathrm{R}_f$ denotes MLPs for dimension reduction. We concatenate the point coordinates, distance offset from the center of the voxel where the point locates, and reflection intensity as the point features $f_p$.

After the voxelization layer, the voxel features are fed into three cascade sparse encoder blocks to obtain sparse semantic features ($F_{s, 1}, F_{s, 2}, F_{s, 3}$). Each sparse encoder block consists of a residual block \cite{he2016deep} with sparse convolutions and an SGFE module developed in \cite{xu2022sparse}. The SGFE module exploits multi-scale sparse projections and attentive scale selection to enhance the voxel-wise features with more geometric guidance and downscales the features' dense resolution by factor 2. 

Also, similar to \cite{xu2022sparse}, we adopt multi-scale sparse supervision to facilitate the learning of semantic context, as shown in Fig. \ref{fig:pipeline}. Specifically, during the training stage, we attach lightweight MLPs as the auxiliary heads after each encoder block to get the semantic predictions of valid voxels. The voxelized semantic labels at different scales are generated by SSC labels according to occupancy grids. Note that point-wise semantic labels are unnecessary since we only apply voxel-wise supervision and compute loss on the valid voxels to avoid unnecessary computation and memory usage. We use lovasz loss \cite{berman2018lovasz} and cross-entropy loss to optimize the semantic branch. The semantic loss $L_s$ is the summation of the loss of each stage, which can be expressed as:
\begin{equation}
    L_s = \sum_{i=1}^{3}{(L_{lovasz,i}+L_{ce, i})}
\end{equation}

Note that the auxiliary heads are removed on the inference stage for efficiency, and our semantic branch only contains 1.45 M parameters.

\textbf{Geometric Representation:} The completion branch takes the occupancy voxels $O_V \in \mathbb{R}^{1\times L\times W \times H}$ generated by the LiDAR point cloud, indicating if voxels are occupied by laser measurements. It outputs multi-scale dense completion features ($F_{c, 1}, F_{c, 2}, F_{c, 3}$) for more geometry details. Since the completion branch aims only to complete the semantic-agnostic scene, i.e., binary completion, we design a shallow architecture with dense 3D convolutions to obtain the geometry details of the scene. As shown in Fig. \ref{fig:rs} (b), the completion branch consists of an input layer and three residual blocks. The input layer is a dense 3D convolution with kernel size $7\times7 \times7$ for a large receptive field, and the residual block is a stack of dense 3D convolutions with kernel size $3\times 3\times 3$. Also, the max-pooling layers are applied before each residual block to downscale the size of the feature map by factor 2. Similar to the semantic branch, deep supervision is used to enhance the multi-scale geometric representation. To this end, we attach MLPs as auxiliary heads after each block to obtain the binary prediction indicating the occupancy of the completed scene. And the training loss $L_c$ for this branch is computed by:
\begin{equation}
    L_c = \sum_{i=1}^{3}{(L_{lovasz,i}+L_{bce, i})}
\end{equation} where $i$ denotes the $i$-th stage of the completion branch and $L_{bce}$ indicates the binary cross-entropy loss. During the inference, the auxiliary heads are removed. And due to the lightweight design (0.31 M parameters), the computational overhead of the completion branch is small (7.93G MACs with input shape $256 \times 256 \times 32$).

\subsection{BEV Fusion Network} \label{Sec.BEV}
Since using dense 3D convolutions to fuse dense 3D feature maps brings a significant overhead on memory usage and slows down the running speed greatly, inspired by the success of BEV perception in 3D object detection and semantic segmentation, we develop a BEV fusion network to aggregate the multi-scale sparse semantic representations ($F_V, F_{s, 1}, F_{s, 2}, F_{s, 3}$) and dense geometric representations ($O_V, F_{c, 1}, F_{c, 2}, F_{c, 3}$) from the BEV.

We first elaborate on the BEV projection of features from semantic/completion branches. For sparse semantic features $F_{s, *}$, we first generate the BEV indices from the voxel indices. Then similar to the voxelization layer in the semantic branch, we use the aggregation function (max function) to aggregate the features with the same BEV index to get the sparse BEV features. Finally, according to the BEV indices and sparse BEV features, we generate the dense BEV features ($F_{s, 0}^b \in \mathbb{R}^{C_0\times H \times W}, F_{s, 1}^b \in \mathbb{R}^{C_1\times (H//2) \times (W//2)}, F_{s, 2}^b \in \mathbb{R}^{C_2\times (H//4) \times (W//4)}, F_{s, 3}^b \in \mathbb{R}^{C_3\times (H//8) \times (W//8)}$). Compared to \cite{yang2021semantic}, which stacks the dense 3D semantic features along the z-axis for BEV features, our used projection method is more efficient and requires less memory overhead. For dense features $F_{c, *}$ from the completion branch, we simply stack the dense 3D features along the z-axis and reduce the feature dimensions with 2D convolutions for dense BEV features ($F_{c, 0}^b, F_{c, 1}^b$, $F_{c, 2}^b$, $F_{c, 3}^b$), which keep the same dimensions as ($F_{s, 0}^b, F_{s, 1}^b$, $F_{s, 2}^b$, $F_{s, 3}^b$).

Similar to \cite{yang2021semantic}, our BEV fusion network is U-Net architecture with 2D convolutions. The encoder consists of an input layer and four residual blocks. Each residual block reduces the resolution size of input features by 2 to keep the same resolution as the semantic/completion features. The concatenation of features $F_{s, 0}^b$ and $F_{c, 0}^b$ is first fed into the input layer and then into the first residual block. Before the next residual block, an Adaptive Representation Fusion (ARF, detailed below) module takes the previous stage's output and semantic/geometric representations at the same scale as inputs and outputs the fused features containing informative semantic context and geometric structure. The decoder upscales the compressed features from the encoder three times by a factor of two at a time through skip connections. And the last convolution of the decoder outputs the SSC prediction $Y \in \mathbb{R}^{((C_n+1)\cdot L)\times H \times W}$, where $C_n$ is the number of semantic classes. The prediction $Y$ is further reshaped as the size of ($(C_n+1)\times L\times H\times W$), representing the semantic occupancy prediction of each voxel of the completed scene. Unlike the semantic/completion branches, we only apply supervision to the final prediction. Both lovasz loss and cross-entropy loss are used to compute the BEV loss $L_{bev}$:
\begin{equation}
    L_{bev} = (L_{lovasz}+L_{ce})
\end{equation}

\begin{figure}[t]
\centering
	\includegraphics[width=0.39\textwidth]{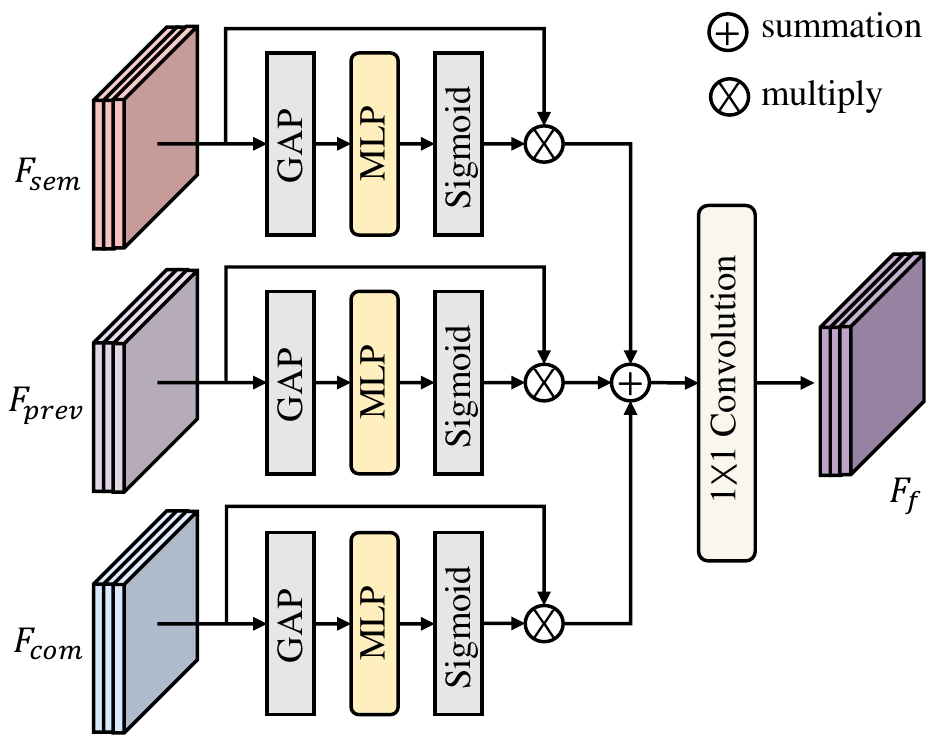}
	\caption{Our designed adaptive representation fusion module. GAP denotes global average pooling.}
	\label{fig:arf}
\end{figure}

\textbf{Adaptive Representation Fusion Module:} Directly concatenating representations from different sources (semantic/completion/BEV branches) similar to SSA-SC implies an equal preference for these representations. However, we usually need selective cues from different sources. To better fuse the semantic and geometric representations, we design an adaptive representation fusion module for the BEV fusion network. Fig. \ref{fig:arf} illustrates the detailed procedure of our ARF module. Let $F_{prev}, F_{sem}, F_{com}$ represent features from the previous stage, features from the semantic branch, and features from the completion branch, respectively. We first compute channel attention for features $F_{prev}/F_{sem}/F_{com}$ to weight the feature channels adaptively. Then the weighted features are summed and passed into a $1\times1$ convolution to obtain the fused features $F_f$. The procedure is formulated as:
\begin{equation}
\begin{aligned}
    F_f = \phi\{\sigma[\mathrm{MLP}(\mathrm{AvgPool}(F_{prev}))]*F_{prev} \\
    +\sigma[\mathrm{MLP}(\mathrm{AvgPool}(F_{sem}))]*F_{sem} \\
    +\sigma[\mathrm{MLP}(\mathrm{AvgPool}(F_{com}))]*F_{com} \}
\end{aligned}
\end{equation} where $\sigma$ denotes the $sigmoid$ function. $\phi$ is the $1\times1$ convolution.

\subsection{Multi-task learning}
We train the whole network end-to-end. The multi-task loss $L_{total}$ is expressed as:
\begin{equation}
    L_{total} = 3\cdot L_{bev}+L_{s}+L_{c}
\end{equation} where $L_{bev}$ is the BEV loss defined in Sec. \ref{Sec.BEV}. $L_{s}$, $L_{c}$ are the semantic loss and completion loss defined in Sec. \ref{Sec.RS}.

\begin{table*}[t]
	\scriptsize
	\setlength{\tabcolsep}{0.005\linewidth}
	\centering
	\caption{Comparison of published methods on the official SemanticKITTI~\cite{behley2019semantickitti} benchmark (hidden test). Our network surpasses all the published methods on completion metrics (IoU) and ranks $2^{rd}$ in terms of the semantic segmentation metrics (mIoU). The speed of LMSCNet, JS3CNet, SSA-SC, and our SSC-RS is tested on the same device(NVIDIA 1080 Ti).} 
	\begin{tabular}{l|c|c c c c c c c c c c c c c c c c c c c|c c}
		\toprule
		Approach 
		& \rotatebox{0}{IoU}
		& \rotatebox{90}{\textcolor{road}{$\blacksquare$} road} 
		& \rotatebox{90}{\textcolor{sidewalk}{$\blacksquare$} sidewalk}
		& \rotatebox{90}{\textcolor{parking}{$\blacksquare$} parking} 
		& \rotatebox{90}{\textcolor{other-ground}{$\blacksquare$} other-ground} 
		& \rotatebox{90}{\textcolor{building}{$\blacksquare$} building} 
		& \rotatebox{90}{\textcolor{car}{$\blacksquare$} car} 
		& \rotatebox{90}{\textcolor{truck}{$\blacksquare$} truck} 
		& \rotatebox{90}{\textcolor{bicycle}{$\blacksquare$} bicycle} 
		& \rotatebox{90}{\textcolor{motorcycle}{$\blacksquare$} motorcycle} 
		& \rotatebox{90}{\textcolor{other-vehicle}{$\blacksquare$} other-vehicles} 
		& \rotatebox{90}{\textcolor{vegetation}{$\blacksquare$} vegetation} 
		& \rotatebox{90}{\textcolor{trunk}{$\blacksquare$} trunk} 
		& \rotatebox{90}{\textcolor{terrain}{$\blacksquare$} terrain} 
		& \rotatebox{90}{\textcolor{person}{$\blacksquare$} person} 
		& \rotatebox{90}{\textcolor{bicyclist}{$\blacksquare$} bicyclist}
		& \rotatebox{90}{\textcolor{motorcyclist}{$\blacksquare$} motorcyclist} 
		& \rotatebox{90}{\textcolor{fence}{$\blacksquare$} fence} 
		& \rotatebox{90}{\textcolor{pole}{$\blacksquare$} pole} 
		& \rotatebox{90}{\textcolor{traffic-sign}{$\blacksquare$} traffic-sign} 
		& \rotatebox{0}{mIoU} 
		& \rotatebox{0}{FPS} \\
		\midrule
		LMSCNet~\cite{roldao2020lmscnet} & 55.3 & 64.0 & 33.1 & 24.9 & 3.2 & 38.7 & 29.5 & 2.5 & 0.0 & 0.0 & 0.1 & 40.5 & 19.0 & 30.8 & 0.0 & 0.0 & 0.0 & 20.5 & 15.7 & 0.5 & 17.0 & 8.5 \\
		LMSCNet-singlescale~\cite{roldao2020lmscnet} & 56.7 & 64.8 & 34.7 & 29.0 & 4.6 & 38.1 & 30.9 & 1.5 & 0.0 & 0.0 & 0.8 & 41.3 & 19.9 & 32.1 & 0.0 & 0.0 & 0.0 & 21.3 & 15.0 & 0.8 & 17.6 & - \\
		Local-DIFs~\cite{rist2021semantic} & 57.7 & 67.9 & 42.9 & \textbf{40.1} & 11.4 & 40.4 & 34.8 & 4.4 & 3.6 & 2.4 & 4.8 & 42.2 & 26.5 & 39.1 & 2.5 & 1.1 & 0.0 & 29.0 & 21.3 & 17.5 & 22.7 & - \\
		JS3C-Net~\cite{yan2021sparse} & 56.6 & 64.7 & 39.9 & 34.9 & 14.1 & 39.4 & 33.3 & \textbf{7.2} & 14.4 & 8.8 & 12.7 & 43.1 & 19.6 & 40.5 & 8.0 & 5.1 & 0.4 & 30.4 & 18.9 & 15.9 & 23.8 & 1.7 \\
		S3CNet~\cite{cheng2021s3cnet} & 45.6 & 42.0 & 22.5 & 17.0 & 7.9 & 52.2 & 31.2 & 6.7 & \textbf{41.5} & \textbf{45.0} & \textbf{16.1} & 39.5 & \textbf{34.0} & 21.2 & \textbf{45.9} & \textbf{35.8} & 1\textbf{6.0} & 3\textbf{1.3} & \textbf{31.0} & \textbf{24.3} & \textbf{29.5} & 1.2\\
        UDNet~\cite{zou2021up} & 59.4 & 62.0 & 35.1 & 28.2 & 9.1 & 39.5 & 33.9 & 3.8 & 0.8 & 0.4 & 4.4 & 40.9 & 23.2 & 32.3 & 0.5 & 0.3 & 0.3 & 24.4 & 18.8 & 13.1 & 19.5 & 13.7 \\
        SSA-SC~\cite{yang2021semantic} & 58.8 & 72.2 & 43.7 & 37.4 & 10.9 & 43.6 & \textbf{36.5} & 5.7 & 13.9 & 4.6 & 7.4 & 43.5 & 25.6 & 41.8 & 4.4 & 2.6 & 0.7 & 30.7 & 14.5 & 6.9 & 23.5 & \textbf{20.0} \\
        \midrule
	  \textbf{SSC-RS(ours)} & \textbf{59.7} & \textbf{73.1} & \textbf{44.4} & 38.6 & \textbf{17.4} & \textbf{44.6} & 36.4 & 5.3 & 10.1 & 5.1 & 11.2 & \textbf{44.1} & 26.0 & \textbf{41.9} & 4.7 & 2.4 & 0.9 & 30.8 & 15.0 & 7.2 & 24.2 & 16.7 \\
		\bottomrule
	\end{tabular}\\
	\label{table:sota}
\end{table*}

\newcommand{\tabincell}[2]{\begin{tabular}{@{}#1@{}}#2\end{tabular}} 
\begin{table*}[t]
	\scriptsize
	\setlength{\tabcolsep}{0.005\linewidth}
	\centering
	\caption{Ablation studies on SemanticKITT validation. BEV, SEM, and COM denote the BEV fusion network, semantic branch, and completion branch. MSS indicates multi-scale supervision (deep supervision) on the semantic/completion branch.} 
	\begin{tabular}{l|ccccc|cc|cc| c c c c c c c c c c c c c c c c c cc}
		\toprule
		Variants & \rotatebox{90}{BEV} & \rotatebox{90}{SEM} & \rotatebox{90}{COM} & \rotatebox{90}{ARF} & \rotatebox{90}{MSS}
	& \rotatebox{0}{IoU}
        & \rotatebox{0}{mIoU}
        & \rotatebox{90}{Precision}
        & \rotatebox{90}{Recall}
		& \rotatebox{90}{\textcolor{road}{$\blacksquare$} road} 
		& \rotatebox{90}{\textcolor{sidewalk}{$\blacksquare$} sidewalk}
		& \rotatebox{90}{\textcolor{parking}{$\blacksquare$} parking} 
		& \rotatebox{90}{\textcolor{other-ground}{$\blacksquare$} other-ground} 
		& \rotatebox{90}{\textcolor{building}{$\blacksquare$} building} 
		& \rotatebox{90}{\textcolor{car}{$\blacksquare$} car} 
		& \rotatebox{90}{\textcolor{truck}{$\blacksquare$} truck} 
		& \rotatebox{90}{\textcolor{bicycle}{$\blacksquare$} bicycle} 
		& \rotatebox{90}{\textcolor{motorcycle}{$\blacksquare$} motorcycle} 
		& \rotatebox{90}{\textcolor{other-vehicle}{$\blacksquare$} other-vehicles} 
		& \rotatebox{90}{\textcolor{vegetation}{$\blacksquare$} vegetation} 
		& \rotatebox{90}{\textcolor{trunk}{$\blacksquare$} trunk} 
		& \rotatebox{90}{\textcolor{terrain}{$\blacksquare$} terrain} 
		& \rotatebox{90}{\textcolor{person}{$\blacksquare$} person} 
		& \rotatebox{90}{\textcolor{bicyclist}{$\blacksquare$} bicyclist}
		& \rotatebox{90}{\textcolor{motorcyclist}{$\blacksquare$} motorcyclist} 
		& \rotatebox{90}{\textcolor{fence}{$\blacksquare$} fence} 
		& \rotatebox{90}{\textcolor{pole}{$\blacksquare$} pole} 
		& \rotatebox{90}{\textcolor{traffic-sign}{$\blacksquare$} traffic-sign} \\
        \midrule
		full model & $\checkmark$ & $\checkmark$ & $\checkmark$ & $\checkmark$ & $\checkmark$ & \textbf{58.6} & \textbf{24.8} & 78.5 & 69.8 & \textbf{73.8} & \textbf{45.3} & 26.6 & 2.1 & 41.0 & \textbf{46.8} & \textbf{41.5} & 1.5 & 6.9 & \textbf{19.8} & \textbf{42.6} & \textbf{22.2} & 50.6 & 6.2 & 1.5 & 0.0 & \textbf{15.8} & \textbf{17.9} & \textbf{4.6} \\ 
        BEV only & $\checkmark$ & $\times$ & $\times$ & $\times$ & $\times$ & 57.1 & 22.0 & 80.6 & 66.3 & 73.5 & 43.5 & 26.1 & \textbf{3.2} & 39.7 & 44.6 & 20.5 & 2.4 & 5.3 & 12.6 & 40.5 & 17.9 & 49.6 & 4.7 & 1.0 & 0.0 & 12.8 & 16.4 & 4.5\\ 
		w/o COM & $\checkmark$ & $\checkmark$ & $\times$ & $\checkmark$ & $\checkmark$ & 58.3 & 23.9 & 80.3 & 68.0 & 72.7 & 45.1 & 18.9 & 0.6 & 40.3 & 47.1 & 37.5 & \textbf{5.4} & \textbf{10.2} & 17.2 & 41.7 & 20.7 & \textbf{51.1} & 6.5 & \textbf{2.7} & 0.0 & 14.4 & 17.4 & 4.6 \\ 
		w/o SEM & $\checkmark$ & $\times$ & $\checkmark$ & $\checkmark$ & $\checkmark$ & 58.1 & 21.9 & 77.9 & 69.6 & 72.5 & 42.3 & 25.1 & 1.3 & 39.3 & 44.9 & 22.2 & 1.0 & 4.6 & 15.4 & 40.0 & 18.5 & 49.9 & 5.5 & 0.9 & 0.0 & 11.1 & 16.4 & 4.4 \\ 
		w/o ARF & $\checkmark$ & $\checkmark$ & $\checkmark$ & $\times$ & $\checkmark$ & 58.4 & 24.4 & 79.8 & 68.6 & 73.4 & 45.0 & \textbf{26.8} & 3.2 & \textbf{41.1} & 46.5 & 36.5 & 5.9 & 8.8 & 18.5 & 41.5 & 20.3 & 50.8 & \textbf{6.8} & 2.0 & 0.0 & 14.5 & 17.2 & 4.2 \\ 
		w/o MSS & $\checkmark$ & $\checkmark$ & $\checkmark$ & $\checkmark$ & $\times$ & 58.1 & 22.7 & 77.4 & 69.9 & 73.5 & 43.5 & 25.8 & 1.7 & 40.4 & 45.7 & 21.7 & 4.1 & 8.2 & 17.3 & 41.3 & 19.7 & 50.4 & 5.3 & 1.2 & 0.0 & 11.2 & 16.2 & 4.1 \\ 
		\bottomrule
	\end{tabular}\\
	\label{table:ab}
\end{table*}

\section{Experiments}
In this section, we introduce the implementation details of the proposed SSC-RS and conduct extensive experiments on the large-scale outdoor scenarios dataset SemanticKITTI \cite{behley2019semantickitti} to show that SSC-RS achieves state-of-the-art performance. Also, we provide visualizations and qualitative analysis to demonstrate the effectiveness of our model. Moreover, ablation studies on semantic/geometric representation, ARF module, and multi-scale supervision are given to validate proposed components.
\subsection{Datasets and Metrics}
\textbf{Datasets} SemanticKITTI \cite{behley2019semantickitti} is based on the KITTI odometry dataset \cite{geiger2012we}, which collects 22 LiDAR sequences with 20 classes in the scenes of autonomous driving using a Velodyne HDL-64 laser scanner. According to the official setting for semantic scene completion, sequences from 00 to 10, except 08 (3834 scans), are for training, sequence 08 (815 scans) is for validation, and the rest (3901 scans) is for testing. The voxelized groud-truth labels with resolution $256\times256\times32$ of train and validation set are provided for the users.

\textbf{Metrics} Following \cite{song2017semantic}, we compute the Intersection-over-Union (IoU) for scene completion (ignoring semantics) and mIoU of $C_n=19$ classes (no ``unlabeled" class) for semantic scene completion as the evaluation protocol. The mIoU is calculated by:
\begin{equation}
    mIoU = \frac{1}{C_n}\sum_{c=1}^{C_n}{\frac{TP_c}{TN_c+FP_c+FN_c}}
\end{equation} where $TP_c$, $TN_c$, $FP_c$, and $FN_c$ denote true positive, true negative, false positive, and false negative for class $c$.
\subsection{Implementation Details}
According to the official protocols, the range $[R_x, R_y, R_z]$ of input point cloud is set $[0\sim51.2m, -25.6\sim25.6m, -2\sim4.4m]$, the voxelization resolution $s$ is 0.2$m$, and the spatial resolution is $(L=256, W=256, H=32)$. The input point cloud is augmented by randomly x-y flipping during the training procedure. And we use Adam optimizer \cite{kingma2014adam} with an initial learning rate of 0.001 ($\beta_1=0.9, \beta_2=0.999$) to train SSC-RS end-to-end. The model is trained for 40 epochs on a single NVIDIA 3090 with batch size 2.

\subsection{Comparison with the state-of-the-art.}
\textbf{Quantitative Results.} We compare with the state-of-the-art on SemanticKITTI test set. We submit the results to the official test server to evaluate the performance of our proposed SSC-RS. Table \ref{table:sota} shows that our SSC-RS achieves the best performance on the completion metric IoU (59.7\%) and ranks $2^{rd}$ in terms of the scene completion metric mIoU (24.2\%). Our SSC-RS also has low latency and runs in real-time (16.7 fps). UDNet \cite{zou2021up}, which adopts dense 3D CNN, has comparable performance on IoU to SSC-RS, while SSC-RS surpasses UDNet by 4.7\% on mIoU and has lower latency. And compared to the semantic segmentation-assistant method SSA-SC \cite{yang2021semantic}, SSC-RS obtains 0.9\% improvement on IoU and 0.7\% on mIoU, which demonstrates the effectiveness of our proposed semantic/geometric representation separation. J3SC-Net \cite{yan2021sparse} attaches semantic scene completion after segmentation. And our SSC-RS also outperforms J3SC-Net by 0.4\% on mIoU, especially 3.1\% on IoU. We notice that our SSC-RS has lower performance on mIoU than S3CNet \cite{cheng2021s3cnet}. We explain that the local geometric loss in S3CNet helps a lot, especially on small objects such as persons, bicycles, and motorcycles, while we don't make a special design on that. Notably, SSC-RS performs better on IoU (14.1\%) and runs $\sim 14 \times$ faster than S3CNet. 

\begin{figure*}[t]
\centering
	\includegraphics[width=0.98\textwidth]{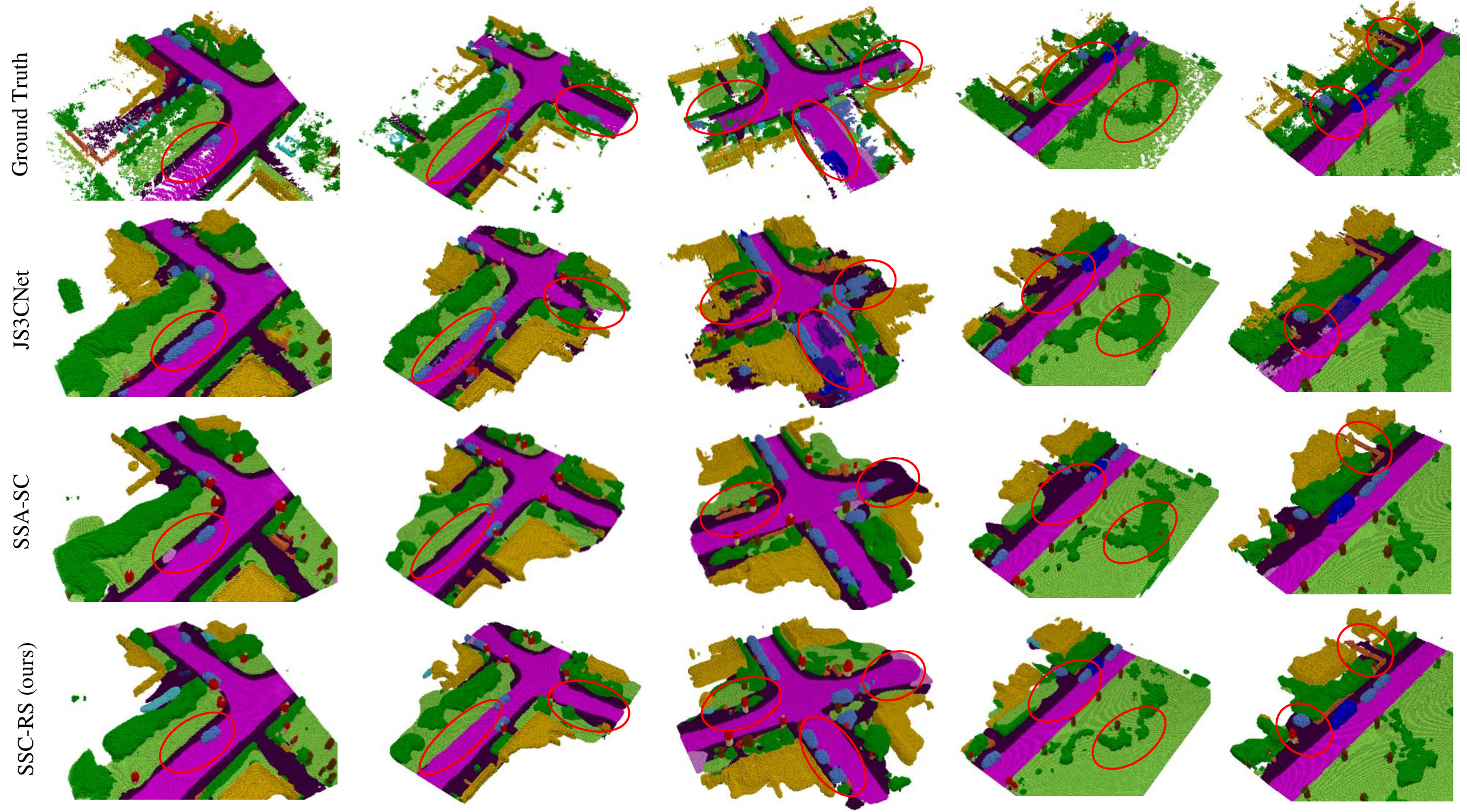}
	\caption{Qualitative results on SemanticKITTI validation set. The visualizations demonstrate that our SSC-RS predicts more accurate SSC results, especially on ``plane" classes and large objects such as cars. (labeled in red circles).}
	\label{fig:visulization}
\end{figure*}

\textbf{Qualitative Results.} We provide the visualizations on SemanticKITTI validation as illustrated in Fig. \ref{fig:visulization}. We also visualize the results of SSA-SC \cite{yang2021semantic} and JS3C-Net \cite{yan2021sparse} for comparison. From Fig \ref{fig:visulization}, we can see that our SSC-RS predicts more accurate SSC results, especially on ``plane" classes and large objects such as cars, consistent with the results in Table. \ref{table:sota}. While we also notice our SSA-SC fails to complete some hard samples, such as small objects, which is also difficult for most methods. We believe that some special designs for local geometry, such as geometry loss in S3CNet \cite{cheng2021s3cnet}, help solve the problem, which also will be our future work.  

\subsection{Ablation Study}
We conduct detailed ablation studies to validate the effectiveness of our proposed components. All experiments are trained with the same training configurations and tested on the SemanticKITTI validation set with a single GPU.

\textbf{Impact of the semantic representation.} We remove the semantic branch to explore the effectiveness of semantic representation. As Table \ref{table:ab} shows, without the semantic branch, the performance drops from 24.8\% to 21.9\% on mIoU (line 1 vs. line 4). Due to the sparse design, the inference speed is still fast (16.7 fps as shown in Table \ref{table:sota}) when equipped with the semantic branch. And compared with the baseline model with the BEV network only, the semantic representation and ARF module improve the performance by 1.9\% on mIoU and 1.2\% on IoU (line 2 vs. line 3). It shows that semantic representation plays a vital role in SSC tasks.

\textbf{Effect of the geometric representation.} We further show the effect of the geometric representation. Line 3 of Table \ref{table:ab} provides the results without the completion branch. The completion branch brings 0.3\% improvement on IoU and 0.9\% gains on mIoU (line 1 vs. line 3) with 0.31 M parameters only. And comparing line 4 with line 2, when fusing with geometric representation using the ARF module, the performance is boosted by 1\% on IoU, demonstrating its effectiveness in improving the accuracy of scene completion. 

\textbf{Ablation study on ARF module.} We remove the ARF module and directly concatenate the features from different sources as the fused features to show the effectiveness of our ARF module. As shown in Table. \ref{table:ab}, 
the ARF module (line 1 vs. line 5) boosts the performance by 0.4\% on mIou and 0.2\% on IoU, which demonstrates the ARF model can select judicious cues from different sources and fuse the representations effectively. 

\textbf{Multi-scale supervision (deep supervision).} Finally, we demonstrate the multi-scale supervision (MSS) on both branches is vital to our SSC-RS. The last line of Table. \ref{table:ab} shows the detailed results. Without multi-scale supervision, the performance drops a lot (2.1\% on mIou, 0.5\% on IoU). It shows that MSS can effectively facilitate the learning procedure of representation separation, which is important for SSC tasks.

\section{Conclusion}
In this paper, we develop SSC-RS to solve outdoor large-scale semantic scene completion from the perspective of representation separation and BEV fusion. Two separate branches with deep supervision are devised to disentangle the learning procedure of semantic/geometric representations explicitly, And the BEV fusion network is designed to fuse the multi-level features effectively and efficiently. Furthermore, an adaptive representation fusion module in the BEV fusion network is proposed to facilitate the fusion procedure. Extensive experiments demonstrate our SSC-RS achieves state-of-the-art performance and runs in real time. We hope our work can provide a new perspective for the SSC community. And in the future, we will focus on local geometry learning to improve the performance on small objects and extend the work to more scenarios, such as indoor and monocular scenes.

\section*{ACKNOWLEDGMENT}
This work was supported by a Grant from the National Natural Science Foundation of China (No. U21A20484).

\bibliographystyle{IEEEtran}
\bibliography{IEEEabrv,ref}


\end{document}